\documentclass[sigconf]{acmart}

\usepackage{amsmath}
\usepackage{booktabs} 
\usepackage{wrapfig}
\usepackage{graphicx}
\usepackage{subfigure}
\usepackage{setspace}

\usepackage{caption}
\begin{document}






%


\title{AxTrain: Hardware-Oriented Neural Network Training for Approximate Inference}

\author{Xin He$^1$, Liu Ke$^1$, Wenyan Lu$^2$, Guihai Yan$^2$, Xuan Zhang$^1$}

\author{$^1$Washington University in St. Louis}

\author{$^2$SKLCA, Institute of Computing Technology, Chinese Academy of Sciences}

\begin{abstract}
The intrinsic error tolerance of neural network (NN) makes approximate computing a promising technique to improve the energy efficiency of NN inference. Conventional approximate computing focuses on balancing the efficiency-accuracy trade-off for existing pre-trained networks, which can lead to suboptimal solutions.
In this paper, we propose AxTrain, a hardware-oriented training framework to facilitate approximate computing for NN inference. Specifically, AxTrain leverages the synergy between two orthogonal methods---one actively searches for a network parameters distribution with high error tolerance, and the other passively learns resilient weights by numerically incorporating the noise distributions of the approximate hardware in the forward pass during the training phase.
Experimental results from various datasets with near-threshold computing and approximation multiplication strategies demonstrate AxTrain's ability to obtain resilient neural network parameters and system energy efficiency improvement.      
\end{abstract}

\copyrightyear{2018} 
\acmYear{2018} 
\setcopyright{acmcopyright}
\acmConference[ISLPED '18]{ISLPED '18: International Symposium on Low Power Electronics and Design}{July 23--25, 2018}{Seattle, WA, USA}
\acmBooktitle{ISLPED '18: ISLPED '18: International Symposium on Low Power Electronics and Design, July 23--25, 2018, Seattle, WA, USA}
\acmPrice{15.00}
\acmDOI{10.1145/3218603.3218643}
\acmISBN{978-1-4503-5704-3/18/07}

\maketitle

\section{Introduction}

An Artificial Neural Network (ANN) is a biologically inspired machine learning model that has been practically demonstrated to deliver superior performance in many recognition, mining, and synthesis (RMS) applications~\cite{RMS}.
The success of ANN can be attributed to innovations across the computing system stack: To achieve higher accuracy, deeper and more complex networks are created along with more advanced training algorithms. To speed up NN training and deployment, powerful parallel computing engines (e.g., GPUs) are designed to accelerate computationally intensive mathematical operations.
Despite the improved performance, energy efficiency still remains a limiting factor when deploying advanced ANNs into edge devices with stringent power budgets. 

A growing body of research has been proposed to tackle energy efficiency from diverse perspectives. Algorithmically, the focus is to simplify neural network (NN) by either using more concise network models (e.g. ResNet~\cite{RESNET} and binary neural networks~\cite{BNN}) or pruning and compressing existing models \cite{DEEPCOMPRESS}. 
From the hardware perspective, efficiency-driven optimizations have been conducted at the architecture, circuit, and device levels. Customized NN accelerators aim at higher energy efficiency, approximate circuits trade accuracy for energy efficiency~\cite{TMSCS,AEYE}, and emerging technologies (e.g. RRAM crossbar) perform low power NN computation in memory~\cite{RRAMTRAIN}.
In this paper, we investigate an auxiliary approach with a focus on network training that can be generally applied to diverse approximate computing techniques. The approach is orthogonally compatible with techniques to improve energy efficiency from other domains.  

Existing approximate computing techniques are confined to exploiting pre-trained NNs, which can result in suboptimal solutions.
Without knowledge of the underlying hardware, NN algorithms optimize only for accuracy under the assumption of ideal hardware implementation, yet they do not consider hardware-specific error tolerance.
Therefore, small noises from approximate hardware may lead to severe network accuracy degradation. Compromises often have to be made to maintain the accuracy target, leading to conservative approximation and failure to exploit all the opportunities for efficiency improvement.

\begin{figure}
\setlength{\abovecaptionskip}{-0.01cm}
\setlength{\belowcaptionskip}{-0.80cm}
\centering
\includegraphics[width=0.38\textwidth]{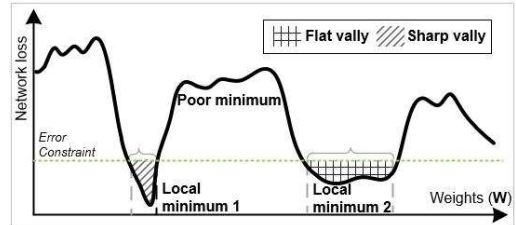}
\caption{Different types of minimums in NN weight space.}
\vspace{-0.3cm}
\label{fig:motivation}
\end{figure}

The key question is how to train a robust neural network that not only achieves high accuracy given ideal hardware assumptions, but is also resilient to noise and errors, so that more aggressive approximation could be applied without severely compromising accuracy.
As Fig.\ref{fig:motivation} illustrates, a conventional training algorithm is dedicated to searching for a ``global'' minimum which has the smallest loss across the weight space, ignoring higher loss in the vicinity of the minimum. Thus perturbations by approximate computing easily results in significant loss, as indicated by ``Local minimum 1''.
Instead of minimizing loss at a single minimum point, our proposed hardware-oriented training seeks a ``near optimal'' minimum where a ``flat'' and ``good enough'' loss surface is preferred and the globally smallest error is not mandatory, as ``Local minimum 2'' depicts.
Thanks to the flat error surface, the NN now exhibits a higher degree of tolerance for approximate computing induced noise. 

In this paper, we propose AxTrain, a hardware-oriented NN training framework for approximate computing.
AxTrain explores two different paths towards high resilience: an active method (AxTrain-act) that explicitly biases the training process to a noise insensitive minimum; and a passive method (AxTrain-pas) that exposes the model of low-level hardware imperfection to the high-level training algorithm for noise tolerance.
AxTrain then leverages the synergy between active and passive methods to facilitate approximate computing.  

In the AxTrain-act method, the innovation is to \textit{guide the training algorithm to improve both network loss and noise resilience directly}.
During training, noise sensitivity is also back propagated along with network loss to the network parameters, and those parameters get updated in order to minimize loss and noise sensitivity.
This solution can be seen as an artificial regularization term to bias the training algorithm towards a high resilience (flat) and accurate (near optimal) minimum, similar to the L2 norm regularization for the over-fitting problem.  

For the AxTrain-pas method, the error tolerance property of the NN is leveraged to reduce side effect from approximate computing.
Rather than training with ideal hardware models, numerical functional models of the approximate hardware are incorporated along the forward pass in the training step, so that the training algorithm can learn the noise distribution of the approximate hardware on its own and \textit{descend to a minimum which is robust to approximate computing}.
Thanks to the knowledge of approximate hardware, the training process experiences different train sets with slightly modified statistical distributions in each epoch, and arrives at a robust model that yields high accuracy with approximate computing.

Finally, to evaluate the effectiveness of the proposed AxTrain framework, we study two popular approximate computing techniques: approximate multiplier and near threshold voltage (NTV) based memory storage \cite{NTC}, because multiplication and parameter storage dominate power consumption in NN accelerators.  

\vspace{-0.4cm}
\section{Related work and Background}
\subsection{Related Work}
Approximate computing is a promising technique for efficiency optimization\cite{JMAO,FUZZY}. Diverse techniques have been explored in prior work that apply approximate computing approaches to improve NN energy efficiency. Minerva~\cite{MINERVA}is an example that uses circuit-level techniques to handle memory error in NN accelerators, employing Razor sampling circuits for fault detection and equipping the weight fetch stage with bit masking and word masking for flipped weights to mitigate bit-flip errors caused by NTV-based weight storage.
Several other prior works on NN accelerators demonstrate the benefit of approximation at the architecture level:
Olivier shows NN accelerators can tolerate transistor-level faults~\cite{OLIVIER}; Zidong \textit{et al.} exploit NN's tolerance for arbitrary approximate multiplier configurations through exhaustive design space exploration~\cite{ZIDONG}.
Recent research proposes more explicit techniques to exploit NN's intrinsic error tolerance and flexibility during training to improve efficiency.
For example, both AxNN and ApproxAnn take neuron criticality into consideration and perform periodical retraining for self-healing \cite{AXNN,APPROXANN}.
AxNN proposes the characterization of neuron criticality first, then the replacement of non-critical neurons with their approximate versions.
To ensure targeted accuracy, iterative retraining is used for error recovery. Inspired by AxNN, ApproxAnn proposes a more reliable way to quantify neuron criticality and adopts iterative heuristics to gain maximum efficiency.

Although AxNN and ApproxAnn both strive to take the advantage of NN's pliable training process to improve energy efficiency, certain limitations in their techniques persist: 1) They require highly configurable hardware where modes of multipliers can be individually adjusted; 2) Due to area and power constraints in large-scale networks with time-multiplexed multipliers, periodic runtime multiplier reconfiguration will inevitably degrade accelerator performance; 3) Approximation is performed on a pre-trained network with hardware-agnostic training, which does not optimize for error tolerance, so the target accuracies in their designs are met with relatively conservative approximations.
All these limitations motivate our AxTrain framework.

\vspace{-0.2cm}
\subsection{Neural Network Preliminary}

\begin{figure}
\setlength{\abovecaptionskip}{-0.01cm}
\setlength{\belowcaptionskip}{-0.50cm}
\centering
\includegraphics[width=0.40\textwidth]{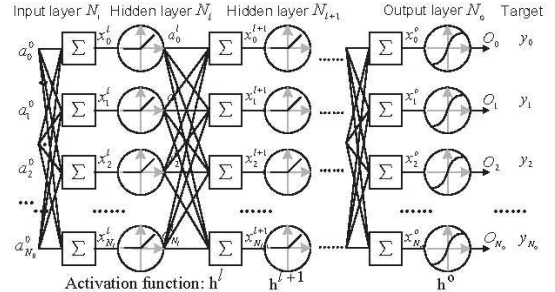}
\caption{A typical neural network topology.}
\vspace{-0.2cm}
\label{fig:neural_network}
\end{figure}
At the architecture level, ANN can be seen as a parallel computing engine which consists of a large number of basic hardware elements, such as multipliers, accumulators, and nonlinear transformation units.
A typical neural network as shown in Fig.\ref{fig:neural_network} consists of an input layer, multiple hidden layers, and an output layer.
During the \textbf{forward pass}, the input layer retrieves inputs, \textbf{$a^{0}$}, of a task sample and directly passes them to the next layer.
To generate activations, $a_{i}$, for each neuron $i$ in a hidden layer, the hidden layer first performs multiplication and accumulation, $\sum_{j=1}^{n} a_j \times w_{ij}$, using activations $a_j$ from the previous layer and network parameters $W$ (including weights and biases), then feeds the intermediate results to a nonlinear transformation $h$, such as Sigmoid ($\frac{1}{1+exp^{-x}}$) for the output layer, ReLu ($max(x,0)$) for hidden layers. This process is repeated layer by layer until the output layer is reached, and the final activations (outputs) from the output layer are generated for regression and classification.

NN training aims at exploring network parameters which minimize the error between network outputs and targets.
To reduce the error, \textbf{backpropagation (BP)} is used to propagate output error from the output layer to the previous layers consecutively and to quantify error contributions from network parameters by taking derivatives of the output error with respect to these parameters.
Then the parameters are updated in a backward pass using stochastic gradient descent to the derivatives to reduce output error.
The mathematical equations for training can be summarized as:

The derivative of the output error with respect to $i$th neuron in layer $l$ is
\begin{equation}
\frac{ \partial{E}}{\partial{x^l_i}} = (\sum_{j=1}^{N_{l+1}} \frac{\partial{E}}{\partial{x^{l+1}_j}} \times w^{l+1}_{ji})\times h'(x^l_i)
\end{equation}
The weights' gradient and updating method are derived as
\begin{equation}
\frac{ \partial{E}}{\partial{w^l_{ji}}} = \frac{\partial{E}}{\partial{x^{l}_j}} \times a^{l-1}_i
\end{equation}
\begin{equation}
w^{l}_{ji} = w^{l}_{ji} - \eta \Delta{w^{l}_{ji}}
\end{equation}
where $\eta$ is the learning rate.
\vspace{-0.3cm}

\section{AxTrain Framework}
In this section, we present the hardware-oriented AxTrain framework that searches for a ``near optimal'' and resilient minimum to facilitate approximate computing and achieve a better tradeoff between inference accuracy and energy efficiency.
Specifically, AxTrain exploits two different methods:
AxTrain-act explicitly regularizes the NN to descend to parameter distributions that are insensitive to noise; and AxTrain-pas intentionally models approximate computing-induced noise in the forward-pass of the training and internalizes the noise distribution in its learned weights.
AxTrain leverages the synergy between the active and passive methods by first training with AxTrain-act to reduce overall sensitivity and then with AxTrain-pas to learn hardware-specific noise.
\vspace{-0.6cm}
\subsection{AxTrain-active Method}
\vspace{-0.2cm}
\subsubsection{Define NN sensitivity-oriented regularization}\hspace*{\fill} \\
\vspace{-0.3cm}

AxTrain-act introduces robustness as an additional regularization term to an NN's cost function to drive NN training. In machine learning, regularization is a process that can introduce prior knowledge to the training process to express preference in the solution.
For example, an L2 regularization term reduces the magnitudes of NN weights and limits NN capacity to prevent over-fitting. Similarly, AxTrain-act defines robustness and incorporates it into the cost function for training, as illustrated below.
\begin{equation}
\vspace{-0.1cm}
E_{tot} = E + \gamma \cdot S(w)
\vspace{-0.1cm}
\end{equation}
where $E$ is the original NN output error. $S(w)$ represents the network sensitivity, and a lower sensitivity suggests higher resilience and more robustness to noise.
We use $\gamma$ as a preference factor for sensitivity.
Based on Eq.4, AxTrain-act minimizes not only network error but also noise sensitivity. 
To reduce the output error, $E$, training algorithm employs backpropagation to evaluate the gradient $\frac{ \partial{E}}{\partial{w^l_{ij}}}$ and update network weights as described in Section 2.

Since the magnitude of $S(w)$ should reflect how output deviations are affected by noisy weights, we define a NN's sensitivity as
\begin{equation}
\vspace{-0.1cm}
S(w) = \sum_k(\sum_{\forall l,ij} |w^l_{ij}||\frac{ \partial{O_k}}{\partial{w^l_{ij}}}|)
\vspace{-0.1cm}
\end{equation}
This definition satisfies four important aspects: 1) We employ absolute values to guarantee that the training process works on worst-case sensitivity reduction, and noises from those sensitive weights cannot cancel out each other to arrive at a smaller $S(w)$. 2) $\frac{\partial{O_k}}{\partial{w^l_{ij}}}$ is the derivative of an output $k$ to a weight $ij$ in layer $l$, which is used to measure the outputs' response with respect to weights perturbation. 3) $|w|$ is also incorporated, since induced noise from the approximate hardware is usually proportional to the magnitude of weight. 4) To minimize heuristic intervention in the optimization process, we capture the total sensitivity by summing across all weights, instead of ranking or partitioning individual weight ~\cite{AXNN}.
Based on this definition, we can infer that a network with small $S(w)$ would behave similarly with and without noise, and hence exhibit better resilience against approximation.
The challenge now is \textit{how to reduce network sensitivity $S(w)$ in training}.
\vspace{-0.4cm}
\subsubsection{Derive gradients}

Inspired by BP and SGD (stochastic gradient descent), we propose to calculate the gradients that measure how the sensitivity changes with respect to the weights and then update the weights accordingly to reduce sensitivity, similar to the conventional BP weight updates for minimizing loss.  

Taking a specific weight $w_{ij}$ as an example, to minimize the sensitivity we should make the update along its negative gradient $\frac{\partial{S(w)}}{\partial{w_{ij}}}$, which can be derived as
\begin{equation}
\begin{split}
\frac{\partial{S(w)}}{\partial{w_{ij}}}&=\frac{\partial{\sum_k(\sum_{ab} |w_{ab}||\frac{ \partial{O_k}}{\partial{w_{ab}}}|)}}{\partial{w_{ij}}}\\
&=\sum_k(sign(w_{ij})|\frac{ \partial{O_k}}{\partial{w_{ij}}}|\\ & + \sum_{ab}(|w_{ab}|sign(\frac{\partial{O_k}}{\partial{w_{ab}}}) \cdot (\frac{\partial^2{O_k}}{\partial{w_{ab}}\partial{w_{ij}}}))
\vspace{-0.3cm}
\end{split}
\end{equation}
The first term, $sign(w_{ij})|\frac{ \partial{O_k}}{\partial{w_{ij}}}|$, is evaluated using BP. 


Evaluation of the second term for all $w$ is complicated because of the second order derivative (Hessian matrix).
Directly calculating the Hessian is a time-consuming process, hence we adopt Pearlmutter's algorithm \cite{HESSIAN} to speed up the computation, since Pearlmutter's algorithm can compute NN's ``Hessian (H) vector (V) product'' in $O(n)$ time simply by \textbf{another round of forward-backward propagation.}
In our case, $|w| \cdot sign(\frac{\partial{O}}{\partial{w}})$ could be denoted as vector $V$, while $\frac{\partial^2{O}}{\partial{w_{ab}}\partial{w_{ij}}}$ as the Hessian matrix $H$ for all parameters.
Pearlmutter's algorithm proposes the R operator which facilitates calculation as
\begin{equation}
R_V\{f(w)\} = \left. \frac{\partial f(w+rV)}{\partial r}\right|_{r=0}
\end{equation}
Hence the second term $V \times H$ of Eq.6 is transformed into $R_V\{\frac{ \partial{O_k}}{\partial{w^l_{ij}}}\}$.
After applying R operator to Eq.2, we have:
\begin{equation}
R_V\{\frac{ \partial{O_k}}{\partial{w^{l+1}_{ij}}}\} = R_V\{ \frac{\partial{O_k}}{\partial{x^{l+1}_i}} \cdot a^{l}_j \} = R_V\{\frac{\partial{O_k}}{\partial{x^{l+1}_i}}\} a^{l}_j +R_V\{ a^{l}_j \}\frac{\partial{O_k}}{\partial{x^{l+1}_i}}
\end{equation}
To compute this equation, we can obtain $\boldsymbol{R_V\{\frac{\partial{O_k}}{\partial{x^{l+1}_i}}\}}$ and $\boldsymbol{R_V\{a^{l}_j\}}$ with a second round propagation as follows: 

1) For the forward pass, the R operator is applied to get $\boldsymbol{R_V\{a_j\}}$:
\begin{equation}
\scriptsize
R_V\{x^{l+1}_j\} = R_V\{\sum_{i=0}^{n} a^{l}_i \cdot w^{l+1}_{ji}\} = \sum_i V^{l+1}_{ji} \cdot a^l_{i} + \sum_i w^{l+1}_{ji} \cdot R_V\{a^l_{i}\} 
\end{equation}
\begin{equation}
\scriptsize
R_V\{a^{l+1}_j\} = R_V\{h^{l+1}(x^{l+1}_j)\} = h^{{(l+1)}\prime}(x^{l+1}_j) \cdot R_V\{x^{l+1}_j\}
\end{equation}
For the input layer, $R_V\{a^{(0)}_j\}=0$. After forward propagation, we can get $R_V\{a^{l}_j\}$; 

2)For the backward pass in the hidden layers to get $\boldsymbol{R_V\{\frac{\partial{O_k}}{\partial{x_i}}\}}$: 
\begin{equation}
\scriptsize
\begin{split}
R_V\{\frac{\partial{O}}{\partial{x^{l}_i}}\} &= R_V\{(\sum_{j=1}^{N_{l+1}} \frac{\partial{O}}{\partial{x^{l+1}_j}} \cdot w^{l+1}_{ji})\cdot h'(x^l_i) \} \\ 
&= h^{''}(x^l_i)R_V\{x^l_i\}(\sum_{j=1}^{N_{l+1}} \frac{\partial{O}}{\partial{x^{l+1}_j}} \cdot w^{l+1}_{ji}) \\
& + h'(x^l_i)(\sum_{j=1}^{N_{l+1}} \frac{\partial{O}}{\partial{x^{l+1}_j}} \cdot V^{l+1}_{ji}) + h'(x^l_i)(\sum_{j=1}^{N_{l+1}} R_V\{\frac{\partial{O}}{\partial{x^{l+1}_j}}\} \cdot w^{l+1}_{ji})
\end{split}
\end{equation}
Here we omit further similar derivation for the output layer.
Once we have $\boldsymbol{R_V\{a_j\}}$ and $\boldsymbol{R_V\{\frac{\partial{O_k}}{\partial{x_i}}\}}$, they can be substituted into Eq.8, and then into Eq.6, and the influence of network weights on sensitivity, $\frac{\partial{S(w)}}{\partial{w_{ij}}}$, can be computed. Note that for AxTrain-act, the training overhead is the time consumed for another round of forward-backward propagation per batch to derive the $\frac{\partial{S(w)}}{\partial{w_{ij}}}$, which does not burden the inference system in an off-line training scenario.  

\subsubsection{Update the preference factor adaptively}\hspace*{\fill} \\
\vspace{-0.1cm}
As defined in Eq.4, AxTrain-act optimizes both the network error and sensitivity, and uses a preference factor $\gamma$ to control the relative magnitude of the sensitivity-related update rate.
A large $\gamma$ may reduce final network accuracy, while a small one could prevent full reduction of sensitivity.
To ensure NN accuracy and convergence, we leverage an adaptive update method for $\gamma$ based on \cite{LAMB}.
Instead of a fixed value, $\gamma$ is updated on a per epoch basis.
A $\Delta \gamma$ is added to $\gamma$ for lower sensitivity if the error in the current epoch is smaller than the weighted sum (0.5, 0.25, 0.125 ...) of training errors in previous epochs or the current error is smaller than a pre-defined accuracy bound. Otherwise, a $\Delta \gamma$ is subtracted to preserve training accuracy.

\vspace{-0.3cm}
\subsection{AxTrain-passive Method}
Different from AxTrain-act, which explicitly optimizes for robustness, AxTrain-pas exposes the nonideality of approximate hardware to the training algorithm by numerically mimicking its inexact operations in the forward propagation. Because of the incorporated hardware knowledge, AxTrain-pas can learn the noise distribution from approximate hardware and implicitly exploit the noise insensitive minimum.         
AxTrain-pas is a hardware-oriented approach which can be generally applied to most approximate techniques in NN accelerators: approximate arithmetic operations \cite{ZIDONG} for neuron calculation and fuzzy memorization \cite{FUZZY} for parameter storage. 

\begin{figure}[t]
\centering
\includegraphics[width=0.30\textwidth]{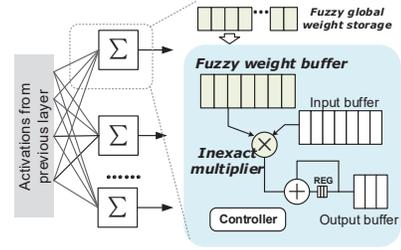}\
\fontsize{8pt}{8pt}\selectfont
\vspace{-0.3cm}
\caption{Approximate computing in neural network accelerators.}
\vspace{-0.5cm}
\label{fig:applyac}
\end{figure}

In neuron calculation, the most computational intensive operations consist of weight activation multiplications and later additions.
Multipliers usually consume higher power and contribute more delays to the critical path than the adders used for accumulations, while the precision of multiplications is relatively less critical than that of additions for NN output accuracy.
All these considerations make multipliers better candidates for approximation, as shown in Fig.\ref{fig:applyac}, where an approximate multiplier is used in a neuron processing element.
Every time a neuron forward calculation, $\sum_{j=1}^{n} a_j \cdot w_{ij}$, is performed, the original accurate multiplications are replaced by their approximate counterparts. 


Power consumption for parameter storage also plays a significant role in NN accelerators, since NNs often consist of thousands of weight parameters.
Fuzzy storage is thus leveraged to trade decreased weight precision for power reduction. Fig.\ref{fig:applyac} also shows fuzzy storage for local and global weights.
To model the effect of approximate computing in the training algorithm, AxTrain-pas models  the noise induced to network weights by fuzzy memorization whenever the weights are retrieved in the forward propagation.
Taking NTV-based fuzzy storage (detailed later) as an example, NTV causes random bit flips, since low supply voltage renders SRAM cells less reliable.
During training, AxTrain-pas models NTV induced flips as stochastic noise \cite{NTC} and injects the noise by randomly flipping the bits in network weights at a certain probability (based on voltage level and technology).
Note that AxTrain-pas applies approximation statically throughout the network.
This policy reduces hardware complexity, such as the support for runtime multiplier reconfiguration and memory mode switching. 

When applying approximate computing in NN, we should first minimize noise from the approximate hardware itself. Taking NTV-based storage as an example, the upper bound of noise in a weight is determined by the binary format used to represent network weights, e.g., the noise magnitude for a sign-bit flip in a fixed-point number corresponds to the maximum value that the fixed point format can represent.
Hence unnecessary high order bits that do not affect accuracy should be eliminated to confine the effect of the noise. Fortunately, most network weights can be easily regularized to concentrated over a range of $10^{-3} \sim 10^{-1}$, so integer bits may not be necessary to represent the weights. In this case the network's activations typically are almost two orders of magnitude larger than the weights, which suggests activations and weights should be represented in different fixed-point formats. Hence, dynamic fixed point representation is used in NN accelerators to maintain network functionality and confine noise~\cite{PRIME}.  

\textbf{Calculate gradients by straight-through estimator.} After augmenting the forward propagation pass with numerical models of approximate computing, a natural question arises: how are the gradients backpropagated through approximate hardware? Given the nonlinear or stochastic nature of approximate hardware, it is hard to analytically compute the precise derivatives across the entire input range for approximate operations.
Inspired by Hinton's lecture (12b) \cite{HINTONLECTURE} and Bengio's work \cite{STOGRA}, we adopt the ``straight-through estimator'' technique in AxTrain-act as below:
\begin{equation}
grad\_in = \left. grad\_out \cdot 1\ \right|_{|grad\_out|<1}
\end{equation}
This BP method directly passes gradients from the outputs of an approximate operator to its inputs, while preventing noise-induced large gradients from disturbing the training algorithm's convergence.
Base on our experimental evaluation, this BP method is effective for AxTrain-pas training. 

\begin{figure}
\setlength{\abovecaptionskip}{-0.01cm}
\setlength{\belowcaptionskip}{-0.4cm}
\centering
\includegraphics[width=0.45\textwidth]{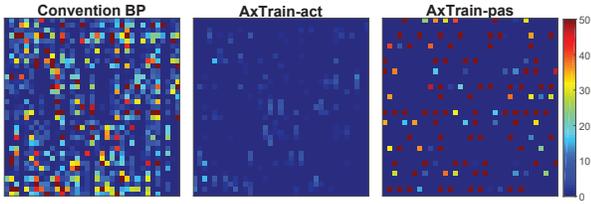}
\caption{Comparison of sensitivity maps between conventional BP, AxTrain-act, and AxTrain-pas.}
\label{fig:sensi}
\end{figure}

We examine the efficacy of AxTrain-act and AxTrain-pas by comparing their weight sensitivity (flatness) with conventional BP.
Fig.\ref{fig:sensi} shows the relative sensitivities of weights from the last (most critical) layer of an multilayer perceptron (MLP) model for the \textit{MNIST digit recognition datasets}, where the deeper blue range indicate less sensitive weights.
AxTrain-pas is trained with approximate multipliers in the most aggressive mode.
Fig.~\ref{fig:sensi} demonstrates the AxTrain-act significantly reduces the sensitivities across all the network weights, while AxTrain-pas implicitly learns noise distribution and selectively reduces the sensitivity for those weights which suffer larger noise from the approximate multiplier.      
\vspace{-0.5cm}









\section{Experimental Methodology}
\begin{figure}
\setlength{\abovecaptionskip}{-0.00cm}
\setlength{\belowcaptionskip}{-0.60cm}
\centering
\includegraphics[width=0.28\textwidth]{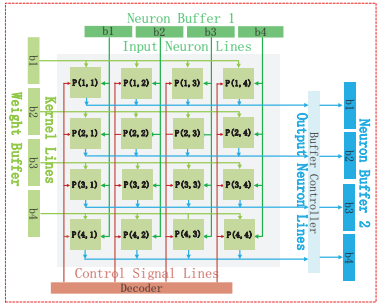}
\caption{FlexFlow Architecture.}
\label{fig:compengine}
\end{figure}
\textbf{NN accelerator architecture.} To evaluate the energy efficiency improvement from approximate computing, we implement a flexible data-driven NN accelerator named ``FlexFlow'' (\cite{FLEXFLOW}) tailored for ANN and shown in Fig.\ref{fig:compengine}.
FlexFlow employs a weight buffer and a neuron buffer for storage, a group of processing engines (PE) for computation, and an instruction decoder for controlling.
To perform neuron calculation, each PE consists of a multiplier, an adder, a neuron local memory, a weight local memory, and a controller.

\textbf{Case studies on two approximate hardware.}
1) \textbf{Approximate multiplier.}
Without loss of generality, to assess the implications of approximate multiplications in NN accelerator, we adopt an existing approximate multiplier for weight-activation multiplication \cite{DRUM}. This design explores the tradeoff between precision and computing efficiency based on changing the effective width $k$ for computation. Generally, in the operands of the multiplier, from the MSB to the LSB only the first nonzero bit and its consecutive $k-1$ bits are retrieved (with the last bit set) for computations. In this way, with a smaller $k$ configuration, the approximate multiplier gains higher energy efficiency at a cost of increased noise. And in the experiment, we adopt four configurations (K1, K2, K3, K4).

2) \textbf{Near threshold voltage storage.}
For fuzzy storage, we leverage NTV supply voltage for SRAM weight storage.
Conventionally, SRAM works as a reliable storage at nominal voltage levels (e.g., 1.1V).
To improve energy efficiency, the SRAM supply voltage can be reduced to the NTV regime at the risk of bit flipping \cite{NTC}.
In this case, the supply voltage can be treated as a knob to tune approximate computing and determine the noise probability. We select two representative knobs (flip rate@voltage: 10\%@400mV, 1\%@660mV, 0.1\%@850mV \cite{NTC}) for each applications, as Table \ref{datasets} depicts.

\textbf{Power evaluation flow.}
To evaluate the power improvement from the approximate multiplier, we first implement approximate hardware using Verilog and then synthesize the design using the Synopsys Design Compiler with the TSMC 65nm library. The power results are gathered using Synopsys PrimeTime. We evaluate the NTV-based storage by CACTI-P\cite{CACTI}. 

\textbf{Training tool and Dataset.}
To evaluate the accuracy of NN, we implement the training algorithm and inference simulator using the PyTorch deep learning framework. The datasets we used are detailed in Table \ref{datasets}. \textit{Breast cancer, Image segmentation, Ionosphere,} and \textit{Satimage} are obtained from the UCI Machine Learning Repository \cite{UCI}, and \textit{MNIST} is a well known dataset for digit classification \cite{MNIST}. We evaluate both the MLP and CNN models for \textit{MNIST}.
For each dataset, 80\% of samples are used for training, while the remaining 20\% are used for testing. In the off-line training, the networks are first trained with AxTrain-act until both the network error and sensitivity cost converge, then tuned with AxTrain-pas for a few more epochs (e.g., 10 epochs for \textit{MNIST}) without hurting the accuracy.

\begin{table}[!tp]
\scriptsize
\centering
\caption{Applications and Parameters.}
\vspace{-0.3CM}
\begin{tabular}{|l|l|l|l|} \hline
\toprule
\textbf{Dataset}&\textbf{Description}&\textbf{NN topology}&\textbf{Agg/Con volt}\\
\midrule
Breast cancer&Diagnose cancer&30,64,64,2&400mV,660mV\\
Image seg&Classify outdoor image&19,64,64,7&660mV,850mV\\
Ionosphere&Identify radar target&34,50,50,2&660mV,850mV\\
Satimage&Classify satellite image&36,64,64,7&660mV,850mV\\
MNIST-MLP&Recognize written digit&784,128,128,10&660mV,850mV\\
MNIST-CNN&CNN for MNIST&LeNet5&660mV,850mV\\
\bottomrule
\end{tabular}
\vspace{-0.4cm}
\label{datasets}
\end{table}
\vspace{-0.2cm}

\section{Experimental results}
\begin{figure}[t]
\setlength{\abovecaptionskip}{-0.01cm}
\setlength{\belowcaptionskip}{-0.20cm}
\centering
\includegraphics[width=0.5\textwidth]{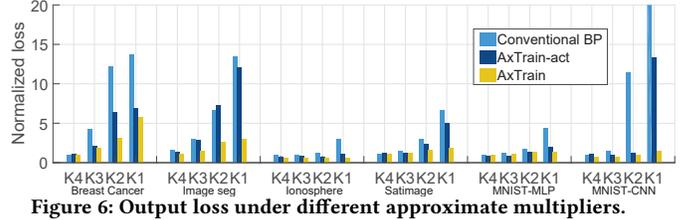}
\caption{Output loss under different approximate multipliers.}
\vspace{-0.3cm}
\label{fig:accuracy_mul}
\end{figure}

\begin{figure}[t]
\small
\centering
\begin{minipage}[t]{0.49\linewidth}
\centering
\includegraphics[width=0.99\textwidth,height=0.77\textwidth]{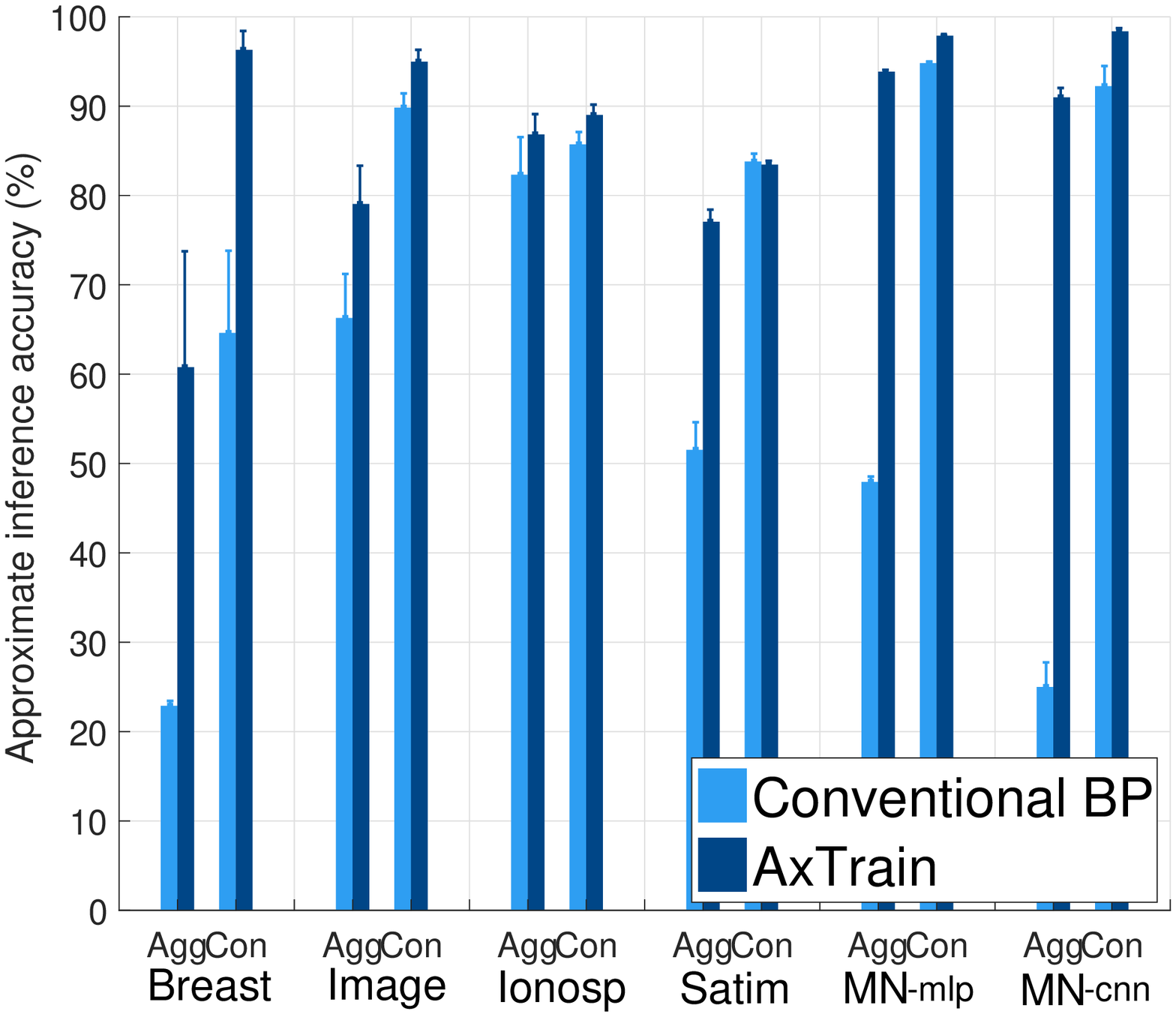}
\vspace{-0.6cm}
\caption{Accuracy comparison between BP and AxTrain under different NTV levels.}
\label{fig:accuracy_ntv}
\end{minipage}
%
\begin{minipage}[t]{0.49\linewidth}
\centering
\includegraphics[width=0.94\textwidth]{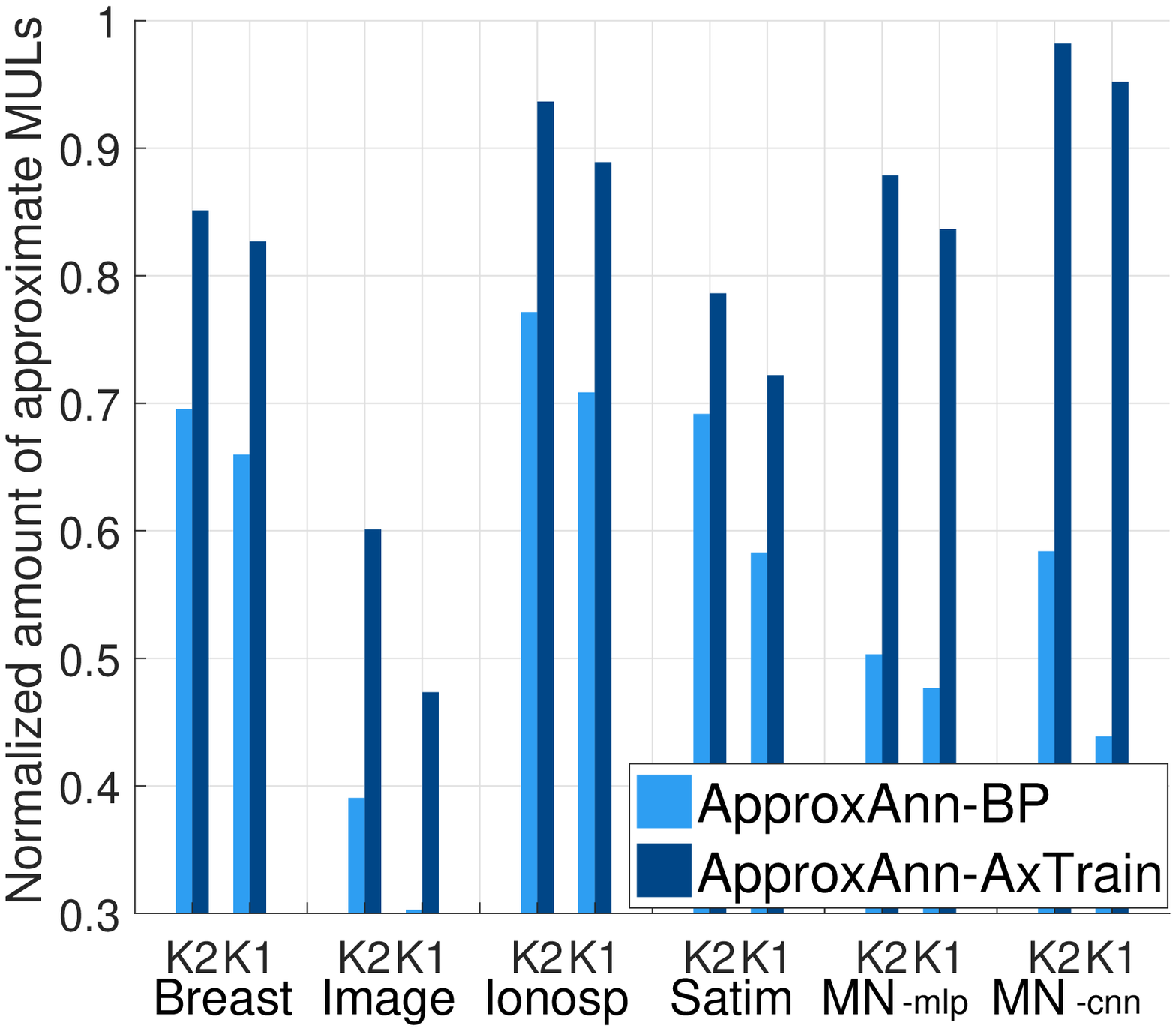}
\vspace{-0.3cm}
\caption{Num of approximate MULs in ApproxAnn-BP and ApproxAnn-AxTrain.}
\label{fig:amount}
\end{minipage}
\vspace{-0.4cm}
\end{figure}


We conduct experiments for six representative applications with four approximate multiplier configurations (K1, K2, K3, K4) and two NTV levels, which include an aggressive (Agg) lower voltage and a conservative (Con) higher voltage, as as Table \ref{datasets} illustrates. 

First, we compare the output error of NN under different approximate multiplier configurations with networks trained by conventional BP, AxTrain-act, and AxTrain, as shown in Fig.\ref{fig:accuracy_mul}.
The errors under different approximate configurations are normalized to original network results with accurate multipliers for each application.
Fig.\ref{fig:accuracy_mul} shows that the network outputs suffer larger error with more aggressive approximation configurations.
Compared with conventional BP scheme, AxTrain-act exhibits higher noise tolerance by reducing error by 40.77\%, 34.56\% and 25.15\% on average for K1, K2 and K3, respectively, while AxTrain further reduces error to 75.61\%, 58.45\%, 37.66\%.
We notice that in a few rare cases (like K4 in \textit{MNIST}), when using multipliers with conservative approximation, AxTrain-act performs slightly better than AxTrain.  
Due to the intrinsic error tolerance of the NN, the accuracy degradation caused by conservative approximate multipliers is quite small, thus the improvement headroom is limited.      


For NTV-based SRAM weight storage, we show the results from fifty runs, since NTV induced bit-flipping is a probabilistic event. Fig.\ref{fig:accuracy_ntv} demonstrates the average accuracies and the deviations. The output accuracies for both aggressive and conservative NTV increase by 32.10\% and 8.163\% on average, compared with conventional BP. This figure also indicates AxTrain reduces the side effects of bit flip in aggressive NTV mode and restores the output quality to a higher level, equals to conventional BP attains in conservative NTV mode.
Note that accuracy is used as the comparison metric instead of network error, because error magnitude for conventional BP in the \textit{MNIST-Agg} case is too large to be properly shown.

For a thorough evaluation of AxTrain, we also implement a recent approach, ApproxAnn, that can be used compatibly with AxTrain-act, thanks to the orthogonality of our training-based approach in supplementing efficient techniques from other domains.
Unlike AxTrain, which uses one approximate configuration throughout the inference, ApproxAnn sets a target accuracy requirement (2\% maximum allowed degradation in this case). It then retrains \textit{a pre-trained BP network} to employ as many approximate multipliers as possible to replace accurate multipliers without exceeding the requirement.
Hence we compare the number of approximate multipliers ApproxAnn could use with pre-trained networks using conventional BP and AxTrain-act, and we show the results for aggressive approximation (K1, K2) in Fig.\ref{fig:amount}.
As expected, NN under less aggressive K2 could always employ a larger number of multipliers than under more aggressive K1.
ApproxAnn with an AxTrain-trained network could use 23.33\% more approximate multipliers on average in the K2 mode and 25.51\% more in the K1 mode than ApproxAnn-BP, which means AxTrain helps ApproxAnn to better exploit the power saving opportunity.
\begin{figure}[t]
\setlength{\abovecaptionskip}{-0.01cm}
\setlength{\belowcaptionskip}{-0.3cm}
\centering
\includegraphics[width=0.5\textwidth]{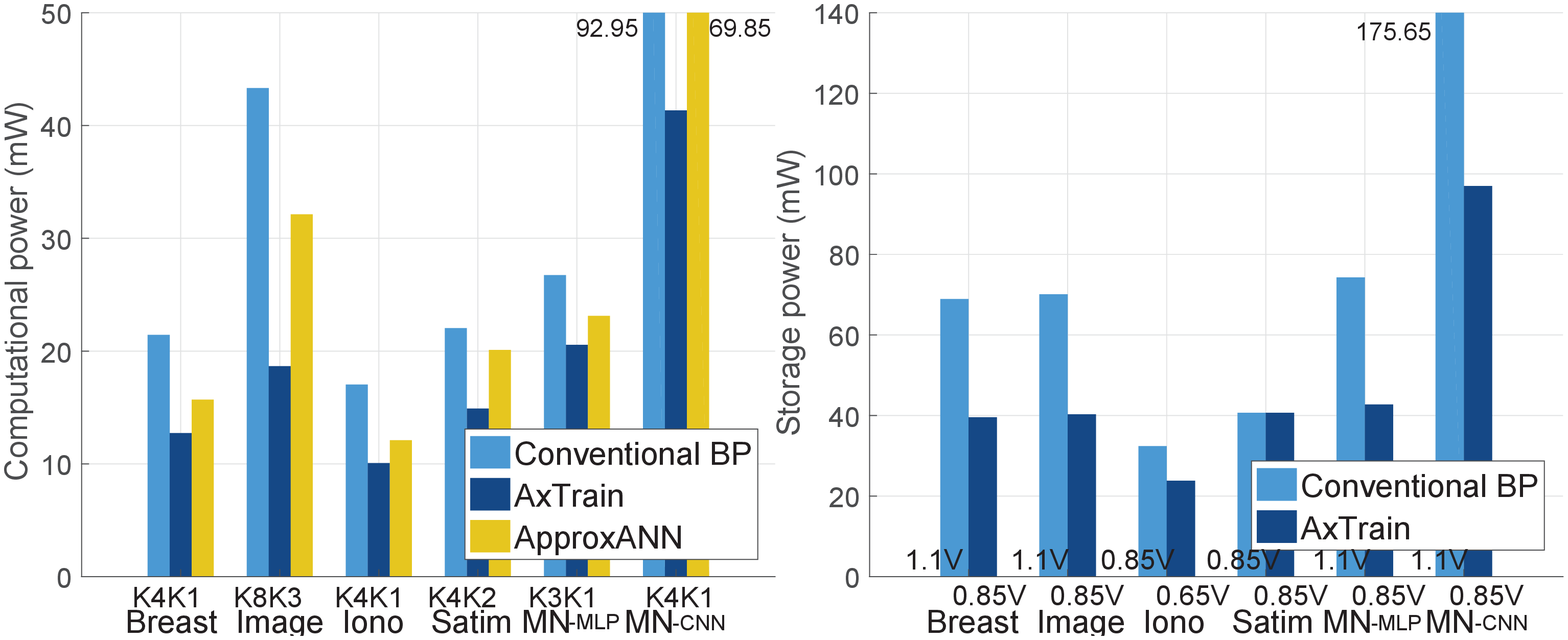}
\vspace{-0.25cm}
\caption{Power consumption under approximate multiplication and NTV based storage.}
\vspace{-0.4cm}
\label{fig:power}
\end{figure}
Finally, to demonstrate the benefit of AxTrain at the system level, we compare the FlexFlow accelerator's lowest power consumption that approximate computing could attain using AxTrain and conventional BP, while keeping a target accuracy (maximum 2\% degradation to accurate implementation), as depicted in Fig.\ref{fig:power}. On the $X$ axis in this figure, we also show the approximation mode that AxTrain and BP apply. This figure shows that AxTrain's higher noise resilience could result in more aggressive approximation, which leads to lower power consumption than conventional BP. To be specific, computational power and storage power are reduced by 41.57\% and 33.14\% on average, correspondingly. We also compare the computational power consumption under approximate multiplier between AxTrain and ApproxANN, and AxTrain requires 25.73\% less power consumption than ApproxANN on average. Notably in the \textit{Satimage} dataset the in NTV storage case, a conservative voltage of 0.85V is enforced to maintain tight accuracy constraint. By relaxing the allowed degradation to 6\%, a 27.01\% power reduction is achieved under 0.66V.



\vspace{-0.4cm}
\section{Conclusion}
\vspace{-0.1cm}
Approximate computing leverages the intrinsic error tolerance of a neural network for improved energy efficiency.The main objective is maintaining good enough accuracy with aggressive approximation. In this paper, we propose the AxTrain framework to optimize NNs for both accuracy and robustness. Using both explicit training and implicit learning, AxTrain reduces NN's sensitivity and improves its resilience against approximation. Experimental results under NTV and approximate multiplier based approximate computing techniques reveal AxTrain could lead to more robust networks than conventional hardware-agnostic training frameworks.
\vspace{-0.3cm}
\begin{acks}
\vspace{-0.1cm}
This work was supported in part by Natural Science Foundation Award \#1657562 and National Natural Science Foundation of China under Grant No. 61572470.
\end{acks}
\vspace{-0.1cm}



%

\bibliographystyle{ieeetr}
{
\fontsize{6.5pt}{6pt}\selectfont 
\bibliography{references}
} 
%
%

\end{document}